\documentclass{article}
\usepackage[utf8]{inputenc}
\usepackage{spconf,amsmath,graphicx}

\usepackage{comment}

\usepackage{algorithm}
\usepackage{algpseudocode}
\usepackage{amsfonts}
\usepackage{multirow}
\usepackage{subcaption}

\begin{document}
\title{Collaborative learning of common latent representations in routinely collected  multivariate ICU physiological signals}
%

%
%
%
\name{Hollan Haule\sthanks{This research was funded in part by an EPSRC Doctoral Training Partnership PhD studentship (to HH) and The Carnegie Trust for the Universities of Scotland (RIG009251) to JE. For the purpose of open access, the author has applied a Creative Commons Attribution (CC BY) licence to any author accepted manuscript version arising from this submission.} \qquad Ian Piper$^{\dagger}$ \qquad Patricia Jones$^{\ddagger}$ \qquad Tsz-Yan Milly Lo$^{\dagger}$ \qquad Javier Escudero$^{*}$ 
}
  
\address{$^{*}$ School of Engineering, Institute for Imaging, Data and Communications (IDCOM), \\
  University of Edinburgh, UK \\
      $^{\dagger}$ Centre of Medical Informatics, Usher Institute, University of Edinburgh, Edinburgh, UK \\
      $^{\ddagger}$ Department of Child Life and Health, University of Edinburgh, Edinburgh, UK}

\maketitle

\begin{abstract}
    In Intensive Care Units (ICU), the abundance of multivariate time series presents an opportunity for machine learning (ML) to enhance patient phenotyping. In contrast to previous research focused on electronic health records (EHR), here we propose an ML approach for phenotyping using routinely collected physiological time series data. Our new algorithm integrates Long Short-Term Memory (LSTM) networks with collaborative filtering concepts to identify common physiological states across patients. Tested on real-world ICU clinical data for intracranial hypertension (IH) detection in patients with brain injury, our method achieved an area under the curve (AUC) of 0.889 and average precision (AP) of 0.725. Moreover, our algorithm outperforms autoencoders in learning more structured latent representations of the physiological signals. These findings highlight the promise of our methodology for patient phenotyping, leveraging routinely collected multivariate time series to improve clinical care practices.
\end{abstract}

%
\section{Introduction}
The wealth of data collected in Intensive Care Units (ICU) offers a significant opportunity to improve disease understanding and patient outcomes via machine learning \cite{Kumar2023ArtificialAgenda}. This is particularly the case for patient phenotyping, where machine learning can identify different groups, or phenotypes, of patients for more effective and personalized treatment. Despite its potential, the exploration of patient phenotyping from time series data has not been extensively studied \cite{Proios2023LeveragingData}.

Several studies have investigated patient phenotype extraction from electronic health records (EHR) using natural language processing \cite{Yang2023MachineReview,Yang2020CombiningRecords,Zeng2019NaturalPhenotyping}. In addition, machine learning techniques involving dimensionality reduction and clustering have been applied for patient phenotyping \cite{Basile2018InformaticsPhenotype}. For example, Rajagopalan \textit{et al.}~\cite{Rajagopalan2022HierarchicalInjury} used hierarchical cluster analysis on hourly physiological measurements, such as heart rate and intracranial pressure, uncovering clusters corresponding to known physiological states in patients with brain injury. However, it is well-recognized that clustering methods typically struggle to scale efficiently with high-dimensional data. Recently, Proios \textit{et al.}~\cite{Proios2023LeveragingData} developed a method integrating graph neural networks with LSTMs, treating patient categorization as a node classification problem using networks of patient similarities and multivariate time series data.


We develop a methodology for learning common states, equivalent to phenotypes shared across patients, from multivariate time series. In contrast to previous research, our approach introduces a novel perspective by focusing on identifying and grouping similar observations within time series data, aiming to pinpoint events that are significant and common among various patients, thereby enhancing the granularity and applicability of patient phenotyping. We hypothesise that such common states are indicative of significant clinical episodes and we create algorithms that leverage LSTM networks and collaborative filtering concepts to learn the states. We then apply this methodology to detect intracranial hypertension (IH), a crucial task in managing patients with severe Traumatic Brain Injury (TBI). Our main contributions are:
\itemsep0em
\begin{enumerate}
\itemsep0em 
    \item A new algorithm for learning and detecting physiological states shared across subjects from multivariate time series, providing a unique way to phenotype patients.
    \item A demonstration of the use of collaborative filtering with LSTM for representation learning.
    \item Learning states related to IH in a clinically relevant, routinely collected minute-by-minute resolution dataset.
    \item Comparisons of the latent representations computed with our method versus a state-of-the-art dimensionality reduction method (variational autoencoder).
\end{enumerate}

\section{Methodology}\label{methodology}
We develop the algorithms in Algorithm~\ref{algorithm} and \ref{inference_algorithm} and conduct a data informatics study on fully anonymized bedside physiological data from pediatric TBI patients in the KidsBrainIT \cite{Lo2018} research databank. In the following, Sections~\ref{principle} to \ref{sec:attention} describe our methodology and Sections~\ref{dataset} to \ref{experiments} cover KidsBrainIT, preprocessing, and experimental design.

\subsection{Core principle and loss function formulation}\label{principle}
We consider a set of $P$ subjects with multivariate recordings:
\begin{equation}
     \textbf{X}:=  \left \{ X_i \mid X_i \in \mathbb{R}^{N_i \times D} \right \}_{i=1}^{P}
\end{equation}
where $X_i$ is a multivariate patient recording of length $N_i$ and $D$ channels. We assume that, for each patient, there exists a latent representation, $Z \in \mathbb{R}^{N_i \times L}$, such that the latent representations during the subsets of $N_i$ when an event of interest happens are similar, as measured by some distance metric.

Using cosine similarity as a distance function, $d(\cdot)$, a similarity matrix between two patients can be defined as 
\begin{equation}
    S_{ij} := d(Z_i,Z_j)
\end{equation}
such that $S_{ij} \in \mathbb{R}^{N_i \times N_j}$ and each element of the matrix represents the similarity between latent vectors in the corresponding timestamps of the two patients. Going forward, we omit the superscripts $i$ and $j$ in $S$, since its implied by the context. In this matrix, similar embeddings will have values approaching $+1$ and dissimilar embeddings will have values approaching $-1$. 

We also define binary label vectors for each patient $\textbf{Y} := \left \{ \textbf{y}_i \mid \textbf{y}_i\in \left \{ -1,1 \right \}^{N_i}\right \}$ , where $1$ indicates timestamps with event of interest and $-1$ otherwise. From this, we define a target matrix, $T$:
\begin{equation}
    T := \textbf{y}_i\textbf{y}_j^{T}
\end{equation}
by taking the outer product between two label vectors. In this matrix, $1$ corresponds to timestamps belonging to similar states, $\left \{ \left ( 1,1 \right ),\left ( -1,-1 \right ) \right \}$ and $-1$ corresponds to timestamps belonging to different states, $\left \{ \left ( 1,-1 \right ),\left ( -1,1 \right ) \right \}$.

To maximize similarity between these embeddings across patients, we use $L_2$ norm between the $T$ and $S$, as our objective function:
\begin{equation}
    loss = \left \| T - S \right \|^{2}.
\end{equation}

This procedure learns a state representation such that similar states occupy neighborhood regions.

\subsection{Training algorithm}
We developed a training procedure in Algorithm~\ref{algorithm}. In an epoch, each patient is compared against the remaining patients in the training set to maximize the similarity of embeddings corresponding to events of interest across patients.

\begin{algorithm}
\caption{Collaborative Training}
\begin{algorithmic}[1]
\State \textbf{Input:} Training data $\textbf{X}_{train}$,$\textbf{Y}_{train}$, Number of epochs $n\_epochs$
\State Initialize model parameters $\theta$: model = $LSTM_{\theta}$
\For{$k = 1$ to $n\_epochs$}
    \For{$i = 1$ to $P$}
        \State $Z_i$ = model($X_i$)
        \For{$j \in \left [ 1:i-1,i+1:P \right ]$ }
            \State $Z_j$ = model($X_j$)
            \State $S = d(Z_i,Z_j)$
            \State $T = \textbf{y}_i\textbf{y}_j^{T}$
            \State $loss = \left \| T - S \right \|^{2}$
            \State Backward pass: $\nabla loss$ with respect to $\theta$
        \EndFor
        \State Update model parameters
    \EndFor
    
\EndFor
\State \textbf{return} Trained model parameters $\theta$
\end{algorithmic}
\label{algorithm}
\end{algorithm}

\subsection{Collaborative Inference by Least Squares}\label{collorativeinf}
During inference, we consider a set of $M$ reference subjects with annotations, $\textbf{R}:=  \left \{ \left ( X_r,\textbf{y}_r \right ) \right \}_{r=1}^{M}$ and a test subject, $X_t$. For each reference subject $\left ( X_r, \textbf{y}_r \right )$ and the test subject $X_t$, we use least squares to compute $\textbf{y}_t$, as
\begin{equation}
    S \approx \textbf{y}_{t} \textbf{y}_{r}^{T}
\end{equation}
\begin{equation}
    \textbf{y}_{t} = S\textbf{y}_{r}\left (  \textbf{y}_{r}^{T}\textbf{y}_{r}\right )^{-1}
\end{equation}
where $S = d(Z_t,Z_r)$ and $Z$ extracted by the model, as described in Section~\ref{sec:model}. The predicted states, $\textbf{y}_{t}$, for each test and reference subject pair are then averaged to give the final prediction, $\textbf{y}_{pred}$. The complete procedure is given by Algorithm~\ref{inference_algorithm}.

\begin{algorithm}
\caption{Collaborative Inference}
\begin{algorithmic}[1]
\State \textbf{Input:} $\textbf{R}$, $X_{t}$
\State Load model parameters $\theta$: model = $LSTM_{\theta}$
\State $\textbf{y}_{pred} \gets 0$
    \For{$r = 1$ to $M$}
        \State $Z_r$ = model($X_r$)
        \State $Z_t$ = model($X_t$)
        \State $S = d(Z_t,Z_r)$
        \State $\textbf{y}_{pred} \gets \textbf{y}_{pred} +  S\textbf{y}_{r}\left (  \textbf{y}_{r}^{T}\textbf{y}_{r}\right )^{-1}$
    \EndFor
\State $\textbf{y}_{pred} \gets \frac{\textbf{y}_{pred}}{M}$
\State $\textbf{return}$  $\textbf{y}_{pred}$
\end{algorithmic}
\label{inference_algorithm}
\end{algorithm}

\subsection{LSTM Model for Extracting Embeddings}\label{sec:model}
We use an LSTM \cite{Hochreiter1997}, trained using Algorithm~\ref{algorithm},  to extract the embeddings from the multivariate patient recordings. LSTM is a popular neural network architecture for sequential modeling tasks. An LSTM maintains a hidden state updated when a new input is observed. We use the hidden state as the embedding for a multivariate sample of a patient recording as
\begin{equation}
\label{LSTM}
\mathbf{z}_t = LSTM(x_t,\mathbf{z}_{t-1}).
\end{equation}

\subsection{Self-Attention and Cross-channel attention}\label{sec:attention}
We include an attention mechanism \cite{Vaswani2017AttentionNeed} in our model, which is a state-of-the-art component in sequential modeling tasks with neural networks. We use standard self-attention mechanism across embeddings in a sequence as
\begin{equation}
    \label{attention}
    {\mathbf{z}}'_i = \sum_{j}^{N}softmax\left(\frac{\mathbf{z}_i^{T}\mathbf{z}_j}{\sqrt{L}}\right)\mathbf{z}_j,
\end{equation}
where $N$ is the length of the signal and $L$ is a scaling factor of size equal to the dimension of the latent space.

In addition, we also consider the suitability of a cross-channel self-attention (CA) to learn the relationships between channels, as
\begin{equation}
    \label{crosschannelattention}
    {\mathbf{c}}'_i = \sum_{j}^{M}softmax\left(\frac{\mathbf{c}_i^{T}\mathbf{c}_j}{\sqrt{K}}\right)\mathbf{c}_j,
\end{equation}
where $M$ is the number of channels and $K$ is a scaling factor. We use a fixed value $K = 60$ since patients have different record lengths. $\mathbf{c} \in \mathbb{R}^{K}$ is a column vector representing a channel in a recording.

\subsection{Dataset description}\label{dataset}
The KidsBrainIT \cite{Lo2018} dataset includes physiological signals such as heart rate (HRT), mean blood pressure (BPm), and mean intracranial pressure (ICPm) from 99 patients collected at minute-by-minute resolution, the clinical standard in most critical care units. It also contains demographic information including age, sex and outcome at 6 and 12 months using Glasgow Outcome Scale. The dataset contains annotations on artifactual samples by an experienced researcher. For additional details about the dataset, see~\cite{Lo2018}.

\subsection{Preprocessing}
We selected 84 patients out of the 99, which had at least 50\% data in mean blood pressure mean, systolic blood pressure, systolic blood pressure, mean intracranial pressure, and heart rate. We use these five signals along with age as the input channels to our model. We remove all artifacts by filtering based on the annotations. We prepare the target IH binary label vector $\textbf{y}$ for each patient, setting a $+1$ when ICPm signal is greater than 15mmHg for at least 48 minutes based on the pressure-time dose curve for children \cite{Guiza2015}, and $-1$ otherwise. Finally, all the physiological signal channels are standardised by subtracting the mean and dividing by the interquartile range of the channel.

\subsection{Experiments}\label{experiments}
We train our model using Algorithm~\ref{algorithm}. We randomly split the patients into the training set (80\%) and the test set (20\%). We use collaborative inference to classify patient states on a sample-by-sample basis in the test set into IH/non-IH. We then evaluate the model using pre-computed labels and report Area Under ROC Curve (AUC) and Average Precision (AP). We repeat the experiments 20 times to generate different training and test sets, since the performance of the approach depends on the variety of patients present in the sets.

We also perform correlation analysis between mean AP per iteration and all patients, to determine the effect of the variety of patients in the reference set on the average AP for an iteration. This is done by creating a table such that each row is iteration result, including columns for mean AP and all patient codes. A patient column is assigned $1$ if the patient is included in the training set and $0$ otherwise. 

We perform an ablation study to determine the importance of cross-channel attention. Specifically, we repeat the experiments described previously 20 times, while cross-channel attention is excluded, and report AUC and AP.

Additionally, we visualize the embeddings by projection into 2D space using t-SNE \cite{VanDerMaaten2008VisualizingT-SNE}. For comparison, we compare the embeddings learned by our technique to those learned by an LSTM-based variational auto-encoder (VAE) \cite{Kingma2014}. The VAE is trained on all patients in the training set, segmented into fixed-length sequences.


\section{Results and discussion}\label{results}
In our study, we employed Algorithm~\ref{algorithm} for training our model and then applied collaborative inference Algorithm~\ref{inference_algorithm} to infer test results in the experiments described in section~\ref{experiments}.

\subsection{Classification performance}
We ran 20 iterations of our experiment. In each iteration, we perform IH classification for each patient in the test set. Table~\ref{tab:combined_table} reports the mean of the metrics, and standard deviation (std), aggregated across the 20 iterations of our experiment with varied training and test sets.

\begin{table}[h]
  \centering
  \caption{Comparison of mean and std of AP and AUC across 20 runs.}
  \begin{tabular}{|c|c|c||c|c|}
  \hline
  \multirow{2}{*}{Metric} & \multicolumn{2}{c||}{With CA} & \multicolumn{2}{c|}{Without CA} \\ \cline{2-5}
  & Mean & Std & Mean & Std \\
  \hline
  AP & 0.698 & 0.110 & 0.725 & 0.074 \\
  \hline
  AUC & 0.836 & 0.106 & 0.889 & 0.042 \\
  \hline
  \end{tabular}
  \label{tab:combined_table}
\end{table}

The effectiveness of our approach in phenotyping is demonstrated by its ability to accurately classify the episodes, with the results showing a clear distinction between IH and non-IH states, well beyond mere chance level. These results are promising but we also noted that, as expected, the performance is intricately tied to the diversity of patients within the training and test sets. Some runs resulted in performance that deviated notably from others.

\begin{figure}[!t]
  \centerline{\includegraphics[width=0.9\columnwidth]{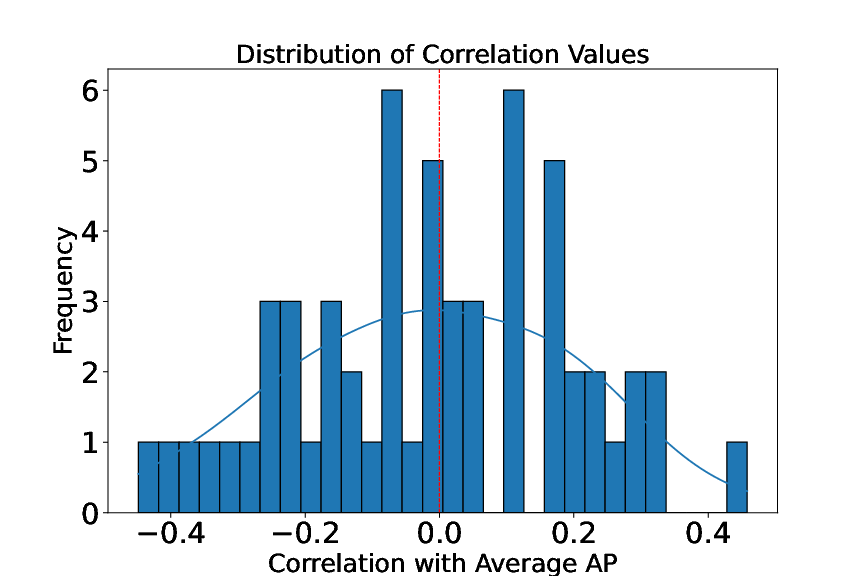}}
  \caption{Distribution of correlation values between the presence of patients in the training set and average AP. This shows that the presence of some patients in the training set strongly affects performance.}
  \label{fig:corr_distribution}
\end{figure}

Fig~\ref{fig:corr_distribution} shows that the presence of some patients in the training set strongly affects the average AP. Such variations in AP across different training dataset configurations underscore the critical importance of patient diversity. Given the collaborative essence of our training and inference methodology, ensuring a well-balanced mix of patient profiles is paramount for achieving consistent and reliable performance. The observed dependence of performance on patient diversity also hints at a potential application of this methodology to uncover patient groups with similar underlying physiology. This, and the exploration of the role of the distance function in our model, will be subject of future research.

\subsection{Cross-channel attention ablation}
 We inspected the role of cross-channel attention in our model. We observe that the addition of cross-channel attention leads to a decrease in overall performance, as shown in Table~\ref{tab:combined_table}. This might indicate that a linear combination of independent channels adds noise, leading to a performance drop.

\subsection{Visualising embeddings with t-SNE}
\begin{figure}[!t]
    \centering
    \begin{subfigure}[b]{0.85\linewidth}
        \includegraphics[width=\linewidth]{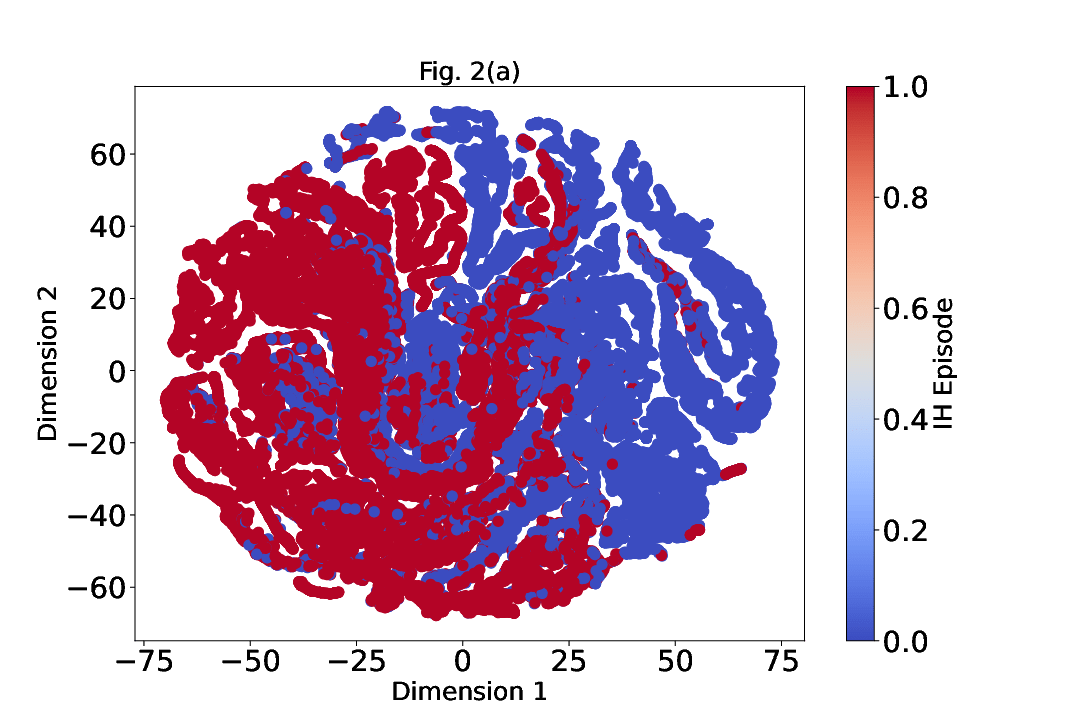}
        \caption{t-SNE 2D projection of patients in a test set using our approach showing more nuanced representation of IH and non-IH states in the patients.}
        \label{fig:embeddings}
    \end{subfigure}
    \hfill 
    \begin{subfigure}[b]{0.85\linewidth}
        \includegraphics[width=\linewidth]{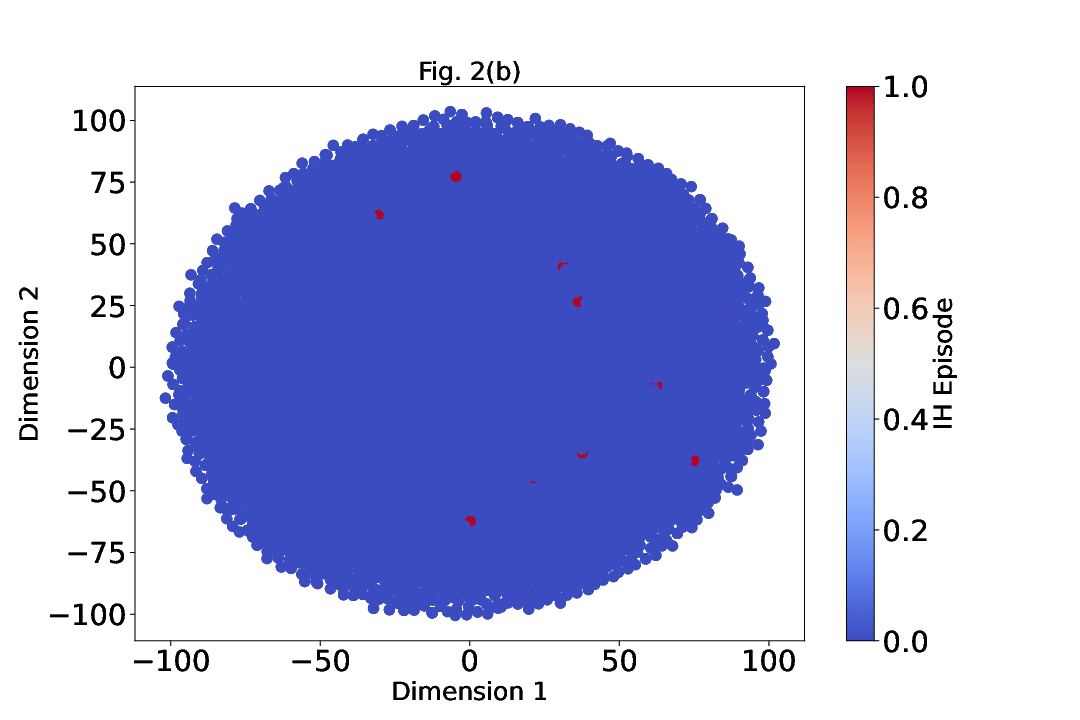}
        \caption{t-SNE 2D projection of patients in a test set using a VAE showing less nuanced representation of IH and non-IH states.}
        \label{fig:embeddings-vae}
    \end{subfigure}
    \caption{Comparative t-SNE 2D projections of patients in a test set.}
\end{figure}

The 2D visualizations, derived from t-SNE projections of our model's embeddings, revealed that IH embeddings are clustered closely together. These patterns, shown in Fig.~\ref{fig:embeddings}, suggest the discriminative power of our approach and its relevance to reveal physiologically meaningul phenotypes. In contrast to the embeddings learned by a VAE, shown in Fig.~\ref{fig:embeddings-vae}, a denser clustering using our approach suggests a more nuanced representation learning, hence improving on state-of-the-art methods for dimensionality reduction.

\section{Conclusion}\label{conclusion}
Our work proposes the use of collaborative filtering concepts to uncover shared latent representations within multiple time series. This approach has demonstrated promising results for phenotyping, demonstrated through the identification of intracranial hypertension states in paediatric TBI patients.

Future endeavors will further analyse the role of each component in the model and validate its adaptability and effectiveness in a variety of challenges, including patient subgroup identification, which represents a major potential application of work. The model may also find applications in representation learning of conditions representing themselves with transient waveforms in physiological signals, such as epilepsy in electroencephalogram brain activity \cite{Jeon2022DeepSpikes} or atrial fibrillation in electrocardiograms \cite{Rizwan2021ALearning}.




\bibliographystyle{IEEEbib}
\bibliography{strings,refs}

\begin{thebibliography}{10}

\bibitem{Kumar2023ArtificialAgenda}
Yogesh Kumar, Apeksha Koul, Ruchi Singla, and Muhammad~Fazal Ijaz,
\newblock ``{Artificial intelligence in disease diagnosis: a systematic literature review, synthesizing framework and future research agenda},''
\newblock {\em Journal of Ambient Intelligence and Humanized Computing}, vol. 14, no. 7, 2023.

\bibitem{Proios2023LeveragingData}
D~Proios, A~Yazdani, A~Bornet, J~Ehrsam, I~Rekik, and D~Teodoro,
\newblock ``{Leveraging patient similarities via graph neural networks to predict phenotypes from temporal data},''
\newblock in {\em 2023 IEEE 10th International Conference on Data Science and Advanced Analytics (DSAA)}, 2023, pp. 1--10.

\bibitem{Yang2023MachineReview}
Siyue Yang, Paul Varghese, Ellen Stephenson, Karen Tu, and Jessica Gronsbell,
\newblock ``{Machine learning approaches for electronic health records phenotyping: a methodical review},''
\newblock {\em Journal of the American Medical Informatics Association : JAMIA}, vol. 30, no. 2, 2023.

\bibitem{Yang2020CombiningRecords}
Zhen Yang, Matthias Dehmer, Olli Yli-Harja, and Frank Emmert-Streib,
\newblock ``{Combining deep learning with token selection for patient phenotyping from electronic health records},''
\newblock {\em Scientific Reports}, vol. 10, no. 1, 2020.

\bibitem{Zeng2019NaturalPhenotyping}
Zexian Zeng, Yu~Deng, Xiaoyu Li, Tristan Naumann, and Yuan Luo,
\newblock ``{Natural Language Processing for EHR-Based Computational Phenotyping},''
\newblock {\em IEEE/ACM Transactions on Computational Biology and Bioinformatics}, vol. 16, no. 1, 2019.

\bibitem{Basile2018InformaticsPhenotype}
Anna~Okula Basile and Marylyn~DeRiggi Ritchie,
\newblock ``{Informatics and machine learning to define the phenotype},''
\newblock {\em Expert Review of Molecular Diagnostics}, vol. 18, no. 3, pp. 219--226, 3 2018.

\bibitem{Rajagopalan2022HierarchicalInjury}
Swarna Rajagopalan, Wesley Baker, Elizabeth Mahanna-Gabrielli, Andrew~William Kofke, and Ramani Balu,
\newblock ``{Hierarchical Cluster Analysis Identifies Distinct Physiological States After Acute Brain Injury},''
\newblock {\em Neurocritical Care}, vol. 36, no. 2, 2022.

\bibitem{Lo2018}
T.~Lo, I.~Piper, B.~Depreitere, G.~Meyfroidt, M.~Poca, J.~Sahuquillo, T.~Durduran, P.~Enblad, P.~Nilsson, A.~Ragauskas, K.~Kiening, K.~Morris, R.~Agbeko, R.~Levin, J.~Weitz, C.~Park, and P.~Davis,
\newblock ``{KidsBrainIT: A new multi-centre, multi-disciplinary, multi-national paediatric brain monitoring collaboration},''
\newblock in {\em Acta Neurochirurgica, Supplementum}, vol. 126. 2018.

\bibitem{Hochreiter1997}
Sepp Hochreiter and Jürgen Schmidhuber,
\newblock ``{Long Short-Term Memory},''
\newblock {\em Neural Computation}, vol. 9, no. 8, 1997.

\bibitem{Vaswani2017AttentionNeed}
Ashish Vaswani, Noam Shazeer, Niki Parmar, Jakob Uszkoreit, Llion Jones, Aidan~N. Gomez, Łukasz Kaiser, and Illia Polosukhin,
\newblock ``{Attention is all you need},''
\newblock in {\em Advances in Neural Information Processing Systems}, 2017, vol. 2017-December.

\bibitem{Guiza2015}
Fabian G{\"{u}}iza, Bart Depreitere, Ian Piper, Giuseppe Citerio, Iain Chambers, Patricia~A. Jones, Tsz Yan~Milly Lo, Per Enblad, Pelle Nillson, Bart Feyen, Philippe Jorens, Andrew Maas, Martin~U. Schuhmann, Rob Donald, Laura Moss, Greet Van~den Berghe, and Geert Meyfroidt,
\newblock ``{Visualizing the pressure and time burden of intracranial hypertension in adult and paediatric traumatic brain injury},''
\newblock {\em Intensive Care Medicine}, vol. 41, no. 6, pp. 1067--1076, 2015.

\bibitem{VanDerMaaten2008VisualizingT-SNE}
Laurens Van Der~Maaten and Geoffrey Hinton,
\newblock ``{Visualizing data using t-SNE},''
\newblock {\em Journal of Machine Learning Research}, vol. 9, 2008.

\bibitem{Kingma2014}
Diederik~P. Kingma and Max Welling,
\newblock ``{Auto-Encoding Variational Bayes},''
\newblock {\em 2nd International Conference on Learning Representations, ICLR 2014}, 12 2013.

\bibitem{Jeon2022DeepSpikes}
Yonghoon Jeon, Yoon~Gi Chung, Taehyun Joo, Hunmin Kim, Hee Hwang, and Ki~Joong Kim,
\newblock ``{Deep Learning-Based Detection of Epileptiform Discharges for Self-Limited Epilepsy With Centrotemporal Spikes},''
\newblock {\em IEEE Transactions on Neural Systems and Rehabilitation Engineering}, vol. 30, 2022.

\bibitem{Rizwan2021ALearning}
Ali Rizwan, Ahmed Zoha, Ismail~Ben Mabrouk, Hani~M. Sabbour, Ameena~Saad Al-Sumaiti, Akram Alomainy, Muhammad~Ali Imran, and Qammer~H. Abbasi,
\newblock ``{A Review on the State of the Art in Atrial Fibrillation Detection Enabled by Machine Learning},''
\newblock {\em IEEE Reviews in Biomedical Engineering}, vol. 14, 2021.

\end{thebibliography}

\end{document}